 \pdfoutput=1 
\documentclass[runningheads]{llncs}
\usepackage{graphicx}

\usepackage{comment}
\usepackage{amsmath,amssymb} 
\usepackage{color}
\usepackage{algorithm2e}
\usepackage{amsmath}
\usepackage{subcaption}
\usepackage{indentfirst}
\usepackage{tikz,pgfplots}
\usepackage{xfrac}

\makeatletter
\newcommand{\printfnsymbol}[1]{%
  \textsuperscript{\@fnsymbol{#1}}%
}
\makeatother

\begin{document}
\pagestyle{headings}
\mainmatter
\def\ECCVSubNumber{7}  

\title{Shift Equivariance in Object Detection} 

\titlerunning{Shift Equivariance in Object Detection}
%
\author{Marco Manfredi\thanks{equal contribution}\orcidID{0000-0002-2618-2493} \and
Yu Wang\printfnsymbol{1}\orcidID{0000-0002-0639-9281}}

\authorrunning{M. Manfredi and Y. Wang}
%
\institute{TomTom, Amsterdam, The Netherlands\\
\email{\{marco.manfredi,yu.wang\}@tomtom.com}}
\maketitle

\begin{abstract}
Robustness to small image translations is a highly desirable property for object detectors. 
However, recent works have shown that CNN-based classifiers are not shift invariant.
It is unclear to what extent this could impact object detection, mainly because of the architectural differences between the two and the dimensionality of the prediction space of modern detectors.

To assess shift equivariance of object detection models end-to-end, in this paper we propose an evaluation metric, built upon a greedy search of the lower and upper bounds of the mean average precision on a shifted image set.
Our new metric shows that modern object detection architectures, no matter if one-stage or two-stage, anchor-based or anchor-free, are sensitive to even one pixel shift to the input images.

Furthermore, we investigate several possible solutions to this problem, both taken from the literature and newly proposed, quantifying the effectiveness of each one with the suggested metric.
Our results indicate that none of these methods can provide full shift equivariance.

Measuring and analyzing the extent of shift variance of different models and the contributions of possible factors, is a first step towards being able to devise methods that mitigate or even leverage such variabilities.
\end{abstract}
\section{Introduction}
Convolutional Neural Networks (CNNs) have achieved impressive results on many computer vision tasks, like classification~\cite{xie2017aggregated}, detection~\cite{cai2018cascade} and segmentation~\cite{chen2017deeplab}. However, in safety critical applications, like autonomous driving or medical imaging, CNNs need to be not just accurate, but also reliable and robust to image perturbations.

The evaluation of the robustness of CNNs has been a very active research field, with several works addressing the impact of image noise~\cite{hendrycks2018benchmarking}, image transformations (translation~\cite{azulay2019deep}, rotation~\cite{engstrom2018rotation}, scale~\cite{singh2018analysis}), object inter-dependencies~\cite{rosenfeld2018elephant,agarwal2020towards} and adversarial attacks~\cite{zhang2019towards}.

Focus of this paper is the robustness of modern object detectors to (small) image translations.
Modern CNN architectures are not shift invariant by design. Commonly used down-sampling approaches, like max pooling, are one cause for shift variance~\cite{zhang2019making}, since these operations are against the classic sampling theorem.

Small transformations can cause a significant drop in classification accuracy for classification models~\cite{engstrom2018rotation}. Furthermore, with the huge receptive field, rather than learning to be shift invariant, modern CNNs filters can derive and exploit absolute spatial location all over the image from zero-padding~\cite{islam2019much,kayhan2020translation}.

Unlike classification, in object detection we need to measure \textit{equivariance} and not \textit{invariance}, since we expect the output of a detector to reflect the translation applied to its input.
\begin{figure}[ht]
\begin{subfigure}{.5\textwidth}
  \centering
  \includegraphics[width=.75\columnwidth]{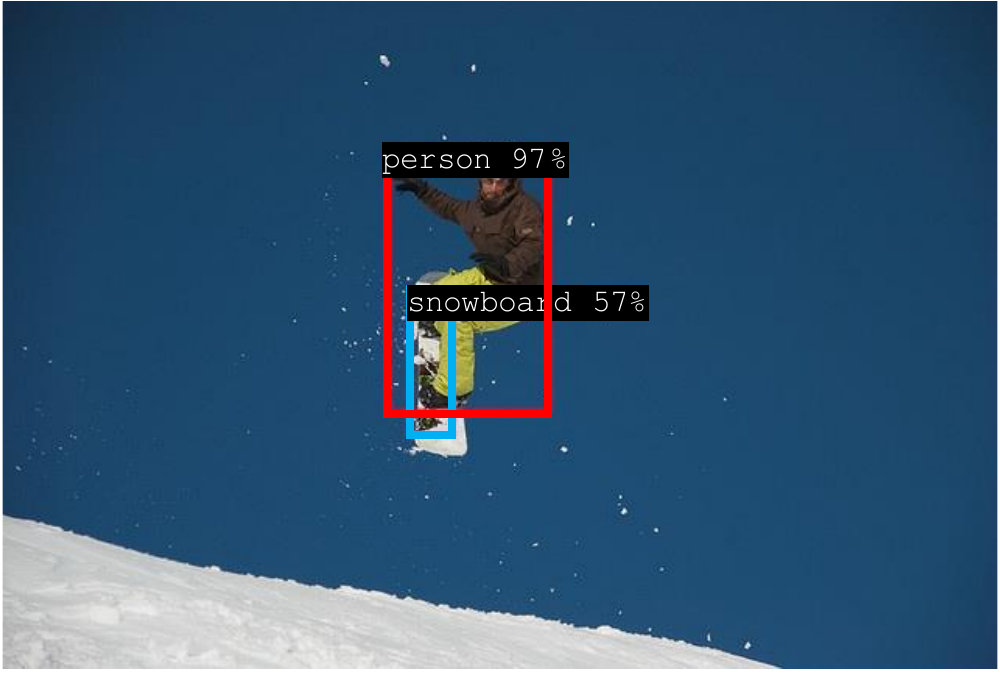}  
\end{subfigure}
\begin{subfigure}{.5\textwidth}
  \centering
  \includegraphics[width=.75\columnwidth]{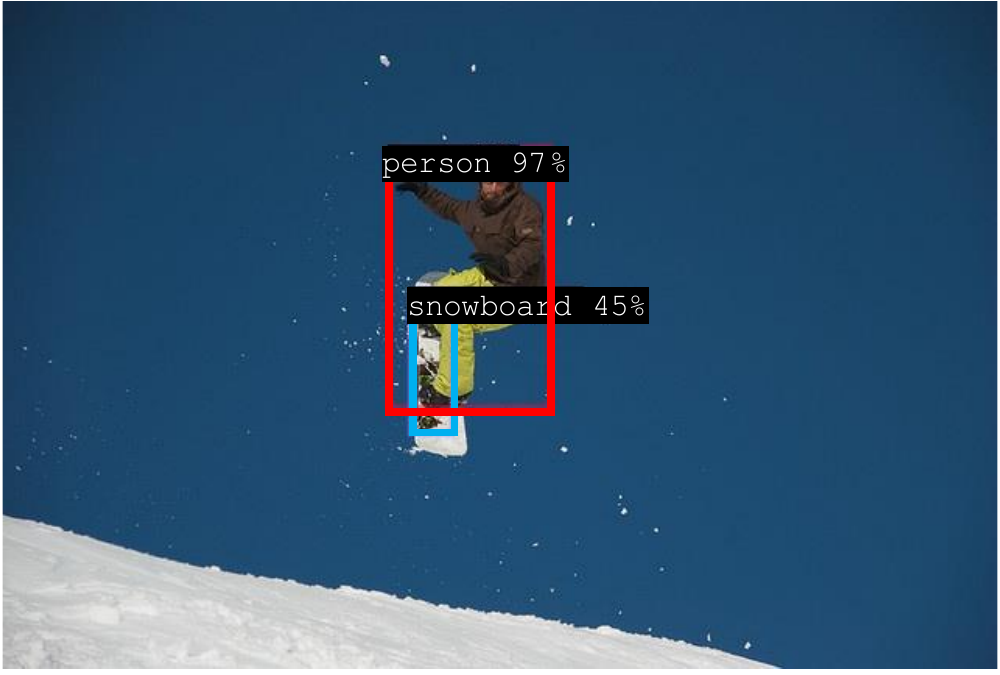}  
\end{subfigure}
\begin{subfigure}{.5\textwidth}
  \centering
  \includegraphics[width=.75\columnwidth]{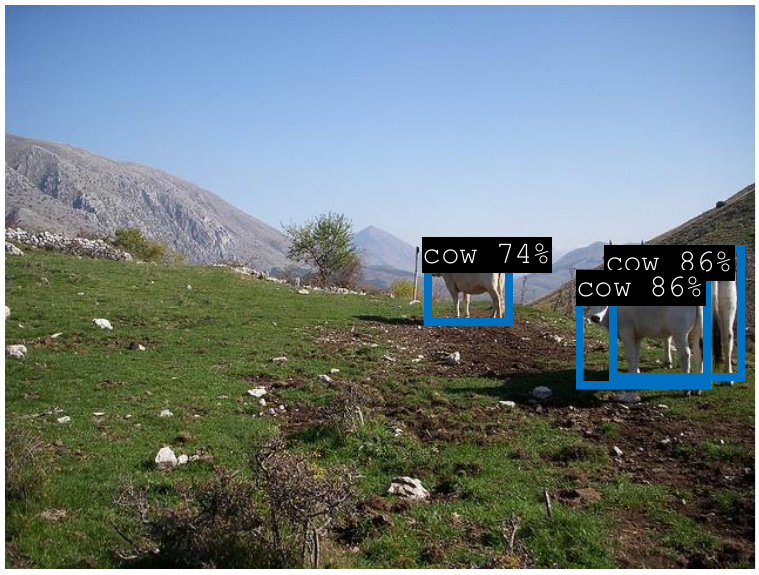}  
\end{subfigure}
\begin{subfigure}{.5\textwidth}
  \centering
  \includegraphics[width=.75\columnwidth]{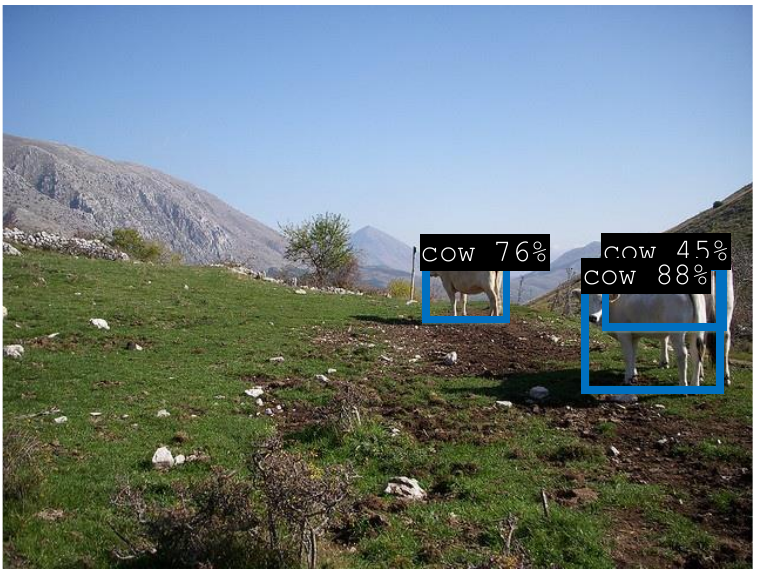}  
\end{subfigure}
\begin{subfigure}{.5\textwidth}
  \centering
  \includegraphics[width=.75\columnwidth]{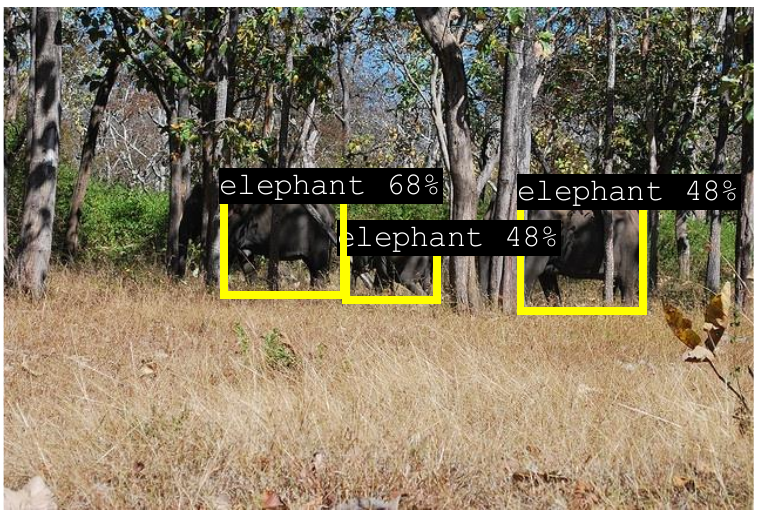}  
  \caption{}
\end{subfigure}
\begin{subfigure}{.5\textwidth}
  \centering
  \includegraphics[width=.75\columnwidth]{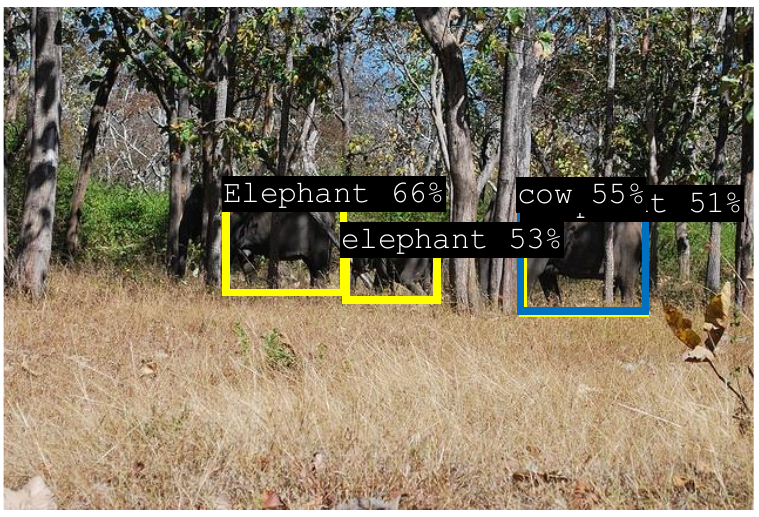}  
  \caption{}
\end{subfigure}
\caption{Qualitative results of shift variance for RetinaNet on the COCO validation set, with confidence threshold set to 0.4 for visualization purposes. The predictions in column a) and b) are computed on images that differ only for a small shift of at maximum one pixel. In the second row we see how both classification and localization of the instances are deeply affected. The third row shows an example of a class swap, where the same instance is predicted as an elephant or a cow.}
\label{fig:qualitative}
\end{figure}
Moreover, each prediction cannot be evaluated independently from all others, given the non trivial assignment between predictions and ground truth instances and the impact of aggregation strategies like non maximum suppression.
To the best of our knowledge, no previous work has proposed a quantitative evaluation of the robustness of detectors to image translations.

We address this gap by designing an evaluation pipeline to quantify shift equivariance of object detection models in an end to end manner.

Previous works also focused on identifying the root causes of the lack of shift invariance, and tried to address them~\cite{hendrycks2019augmix,zhang2019making,sundaramoorthi2019translation}.
However, the impact that these factors might have on object detectors has not been investigated yet. We believe this has non-trivial implications, since detection networks extend classification backbones with additional modules like regression/classification heads (both anchor-based and anchor-free), Region Proposals Networks or ROI Pooling/Align modules.
We therefore conduct experiments to dissect translation equivariance on object detection networks: anti-aliasing the intermediate features, densifying the output resolution, using dataset which is free of photographer's bias and test-time augmentation.

Our contributions are:
\begin{enumerate}
    \item We design an evaluation pipeline and a metric, AP variations ($\Delta AP$), specifically targeted to shift equivariance in object detectors;
    \item We test several modern object detectors with the proposed metric and we report a severe lack of robustness to translations as small as one pixel, on two different datasets;
    \item We measure the effectiveness of different techniques to enhance robustness, and show that they provide only a partial answer to the problem.
\end{enumerate}

\section{Related Work}

{\bfseries Network robustness}
Convolutional neural networks are reported to have brittle robustness to perturbations. Adversarial attacks~\cite{goodfellow2014explaining,kang2019testing}, which are intentionally designed noises to fool the network to make wrong predictions, are used to investigate network failures. Recent work has found that models are also brittle to less extreme perturbations that are not adversarially constructed, such as image quality distortions, i.e. blur, noise, contrast and compression~\cite{dodge2016understanding}. 
CNNs are reported to not be robust to scale variations~\cite{singh2018analysis} and highly sensitive to pose perturbations~\cite{Alcorn_2019_CVPR}.  Networks fail to even small rotations and translations~\cite{azulay2019deep,engstrom2018rotation,zhang2019making}. 
Imperceptible variations and natural transformations across consecutive video frames would induce instability for networks as well~\cite{azulay2019deep,gu2019using}.

There are also works on the robustness of object detectors. Image corruptions such as pixel noise, blur, varying weather conditions lead to significant performance drops in object detection models~\cite{michaelis2019benchmarking,von2019simulating}. Object detectors are not robust against scale variations~\cite{singh2018analysis}. Replacing image sub-regions with another sub-image is shown to have non-local impact on object detection, which may occasionally make other objects undetected or misclassified~\cite{rosenfeld2018elephant}.

{\bfseries Network robustness evaluation} The robustness is often evaluated for classification networks in terms of classification label swap or mean absolute score changes. For example attack success rates (changes of top-1 prediction) across several distortion sizes are used to measure robustness with a broader threat model and diverse differentiable attacks \cite{kang2019testing}. Azulay et al. quantify both top-1 and mean average score changes of the classifier to small translations or re-scaling of the input images \cite{azulay2019deep}.
\cite{hendrycks2018benchmarking} quantifies the classifiers mean and relative error rate on corruptions with different severity level to show the networks robustness against a variety of common corruptions.
To measure robustness against geometric transformations, Manitest defines the invariance as the minimal geodesic distance between the identity transformation and a transformation in Lie group $\tau$ which is sufficient to change the predicted label of the classifier~\cite{fawzi2015manitest}.
In our paper, we focus on evaluating shift equivariance of object detectors, considering both classification and localization performance. 

{\bfseries Approaches to improve shift invariance} One approach to improve the robustness of the model is to apply data-augmentation such as random cropping to make the network learn to be invariant to shifts~\cite{hendrycks2019augmix}. However,~\cite{azulay2019deep} has shown that data augmentation teaches the network to be invariant to translations but only for images that are visually similar to typical images seen during training, i.e. images that obey the photographer’s bias.
Another way to mitigate shift variance is to apply anti-aliasing operations by blurring the representations before downsampling~\cite{zhang2019making}, as often done in signal processing. Although~\cite{zhang2019making} observed increased accuracy in ImageNet classification,~\cite{azulay2019deep} has shown that blurring would degrade model performance, especially in datasets that obey the photographer’s bias.
GaussNet CNN architecture, which represents convolutional kernels with
an orthogonal Gauss-Hermite basis whose basis coefficients are learned in convolution layers without a sub-sampling layer, leads to fully translation invariant representations that keeps the number of parameters in kernels across layers constant~\cite{sundaramoorthi2019translation}. 
In our paper, we also investigate several solutions to alleviate translation variance, focusing the analysis on object detection models.

\section{Measuring Translation Equivariance for Object Detectors}
A good measure of translation equivariance should capture how the output of a detector changes when a shift is applied to the input image. The ideal detector would output the same predictions, but shifted. 

In order to measure translation equivariance we need to define: an experimental setting to generate shifted images and a metric to measure the amount of change in the detector output. For the experimental setting, we follow other works on shift invariance for classification networks~\cite{azulay2019deep}, please refer to our experimental section for further details.

The shift invariance metrics for classification measure the changes in per-image class scores, for example reporting the gap between the highest and lowest class scores predicted by the network on shifted images.
In object detection, it is possible to compute these metrics at the granularity of single instances, evaluating the probability score variations of each ground truth object.
However, object detection is a more complex task than classification, and the metrics to measure robustness to translation should be tailored specifically for it.
We believe a good metric for translation equivariance in object detection should:
i) consider all predicted boxes, also the ones on background regions (i.e. potential false positives) and ii) include localization accuracy in its formulation.
In general, this robustness metric should be a clear indicator of the performance variations we might get when perturbing the input with small translations.

This motivates the introduction of AP variations as our proposed indicator of translation equivariance in object detectors.

\subsection{AP Variations}
Average precision is the most common way of evaluating object detectors~\cite{lin2014microsoft}. A natural formulation for a shift equivariance metric is thus the variations of average precision over a (shifted) dataset. This metric captures the performance gap between the worst and best case.

Given a set $\mathbf{X} = ({X_1, X_2, \cdots, X_N})$ of $N$ images and a maximum shift range $M$ (e.g. 1 pixel), we compute a shifted set, where each image $X_i$ is present multiple times, slightly shifted horizontally and vertically, $\mathbf{X}_i^{\Delta} = \{X_i^{\delta}  \mid {\delta} \in \{(0,0), (0,1), \cdots, (M,M) \}\}$.

The AP variation over the shifted dataset is defined as the difference between the highest and lowest achievable APs: $\Delta{AP} = AP_{best}-AP_{worst}$.
As an example, $AP_{best}$ is the maximum AP achievable by selecting the best shift $\delta_i^{*}$ for each image. In other words, for each image the shift that contributes to the highest overall AP is selected.
In practice, we use $AP_{50}$ as the metric, that is, using 0.5 as the IoU threshold to define a positive match between a prediction and a ground truth object.
\subsubsection{Greedy Approximation of AP variations}
Average precision is computed on the entire dataset by sorting the predictions of all images by confidence score, and evaluating the ranking of true positives and false positives. This means that we can't find the best (and worst) shift for each image independently from all other images.
In theory, finding the best AP would involve computing the AP of all combinations of shifts across images.
This is computationally infeasible, since the number of combinations would be $(M +1)^{2N}$. For the validation set of COCO, and a maximum shift of 1 pixel, that would be $2^{10000}$ total AP computations.

We propose an approximation to this optimal solution using a greedy algorithm that iteratively finds the best shift $\delta_i^{*}$ for each sample $X_i$ while keeping fixed the solutions of all other samples $X_{j, j\neq i}$.
Algorithm~\ref{alg:greedy} details the procedure to compute $AP_{best}$, by switching $\arg\max$ with $\arg\min$ we can compute $AP_{worst}$.

The number of AP computations becomes now tractable and is equal to $K \times (M+1)^2 \times N$.
In practice, setting the number of iterations $K=1$ is enough for the algorithm to converge to a stable solution, since in our experiments further iterations contributed for less than 1\% to the total AP difference.

\begin{algorithm}
\SetAlgoLined
\textbf{Input:}

 $\mathbf{X}$: the set of images
 
 $\Delta = (\delta_{0}, \delta_{1}, \cdots)$: set of shifts per samples
 
 $K$: number of iterations
 
 
 $e$: evaluation function
 
 $f$: detector inference function
 
 Initialize  $\delta_i^* = \delta_{0}$  (for each sample $i$, shift (0,0) is selected)
 
 \For{$k$ in $K$}
 {
  \For{$i$ in $N$}
  {
    $AP$ = [ ]
    
    \For{each $\delta$ in $\Delta$}
    {
        $AP_{\delta} = e(f(X_i^{\delta} \cup \mathbf{\tilde{X}}))$,  $\mathbf{\tilde{X}} = \{X_j^{\delta_j^*} \mid j \neq i, j \in \{1,\cdots, N\}\}$
        
        $AP$.append($AP_{\delta}$) 
    }

   \#update the best performing shift for sample $X_i$
   
   $\delta_{i}^{*} = \underset{\delta \in \Delta}{\arg\max}$ $AP$
  }
 }
 
 $AP_{best} = e(f(\mathbf{X^*}))$, $\mathbf{X^*} = \{X_i^{\delta_i^*} \mid i \in \{1,\cdots, N\}\}$
 
 \caption{Greedy algorithm to approximate the best Average Precision by iteratively selecting the best shift for each sample.}
 \label{alg:greedy}
\end{algorithm}

\section{Experiments}
We evaluate shift equivariance for three common object detectors: RetinaNet~\cite{lin2017focal} (one-stage, anchor-based), CenterNet~\cite{ren2015faster} (one-stage, anchor-free) and FasterRCNN~\cite{zhou2019objects} (two-stage) on the COCO object detection dataset~\cite{lin2014microsoft}.
For RetinaNet and FasterRCNN, we use the pretrained models with ResNet-50 backbone available in the detectron2 framework~\cite{wu2019detectron2}, for CenterNet we use the pretrained model with ResNet-101 with deformable convolutions from the official repository~\footnote{github.com/xingyizhou/CenterNet}. 

\begin{figure}[ht]
\captionsetup[subfigure]{labelformat=empty}
\begin{subfigure}{.24\textwidth}
  \centering
  \includegraphics[width=.8\linewidth]{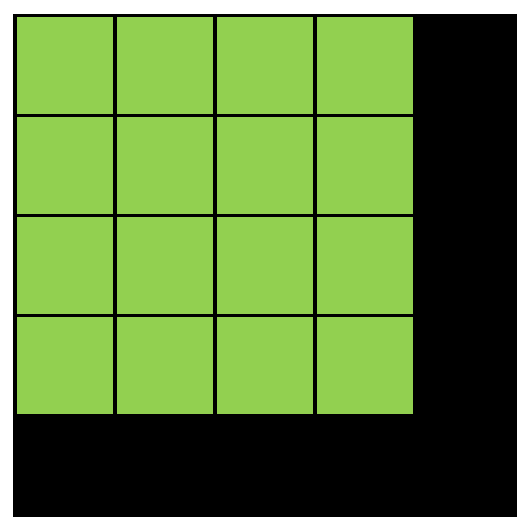}  
  \caption{shift (0,0)}
  \label{fig:sub-first}
\end{subfigure}
\begin{subfigure}{.24\textwidth}
  \centering
  \includegraphics[width=.8\linewidth]{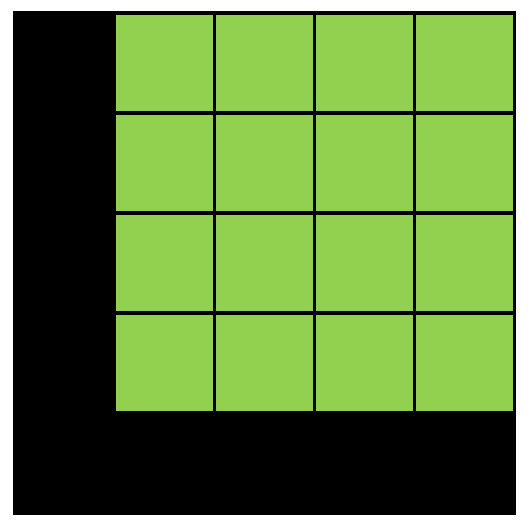}  
  \caption{shift (0,1)}
  \label{fig:sub-second}
\end{subfigure}
\begin{subfigure}{.24\textwidth}
  \centering
  \includegraphics[width=.8\linewidth]{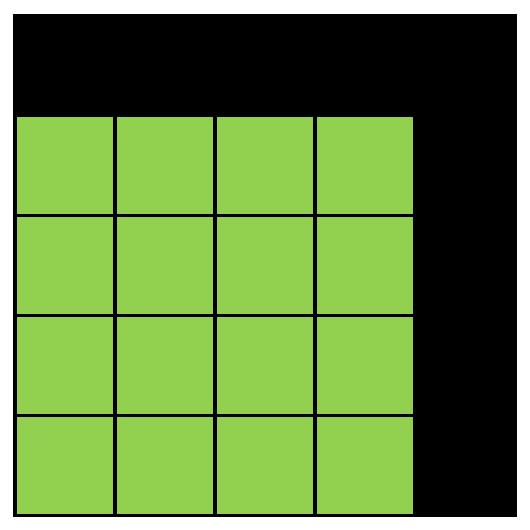}  
  \caption{shift (1,0)}
  \label{fig:sub-first}
\end{subfigure}
\begin{subfigure}{.24\textwidth}
  \centering
  \includegraphics[width=.8\linewidth]{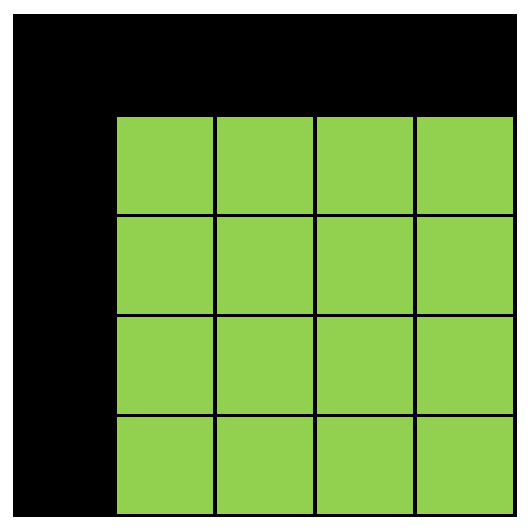}  
  \caption{shift (1,1)}
  \label{fig:sub-first}
\end{subfigure}
\caption{Creating the dataset for shift equivariance evaluation. An example 4$\times$4 image (in green) is embedded into a black image at different locations (maximum shift=1) to create the 4 translated versions.}
\label{fig:translated_images}
\end{figure}

In order to evaluate shift equivariance, we create a shifted validation set, where each image appears multiple times, shifted horizontally and vertically. We follow the experimental setting proposed in~\cite{azulay2019deep}, where each image is embedded in a black background image at different locations. This setting prevents to crop out any context from the image, ensuring the only difference is a small translation, see Figure~\ref{fig:translated_images}. This setting is particularly appropriate for object detection, since objects are located anywhere in the image, and cropping even a small portion of it might imply removing relevant object parts.

Given a maximum shift, we compute predictions for all shifted images and the best and worst overall APs using the proposed method. $\Delta AP$ is obtained as the simple difference of the two. In Table~\ref{table:ap-variations-coco}, the results for the entire COCO validation set are reported for a maximum shift of one pixel.

All three methods show a surprising lack of shift equivariance, with CenterNet being the least robust. We show some qualitative examples in Figure~\ref{fig:qualitative}.
CenterNet is the only method of the three that does not use NMS to aggregate overlapping predictions. Instead, it uses a max-pooling layer to extract local peaks in the keypoint heatmap. This is a simpler implementation of NMS that might be less effective to retain the best boxes.

The $\Delta APs$ are comparable to the performance gaps between these methods, meaning that a one pixel shift is enough to change what we might consider state-of-the-art performance.
CenterNet is the only method trained with extensive data augmentations, including random cropping. Its poor performance in terms of shift equivariance confirms the conclusions from~\cite{azulay2019deep}: data augmentation is not enough to improve robustness. 

\begin{table}
\centering
\begin{tabular}{|l|l|l|l|l|l|l|}
\hline
\textbf{method}       & $\mathbf{AP}$ & $\mathbf{\sfrac{worst}{best}}$ $\mathbf{AP}$  & $\mathbf{\Delta AP}$ & $\mathbf{AP_{50}}$ & $\mathbf{\sfrac{worst}{best}}$ $\mathbf{AP_{50}}$  & $\mathbf{\Delta AP_{50}}$ \\ \hline
RetinaNet   &  36.5 & 35.3/37.5   & 2.2  & 56.7 & 53.9/59.0 & 5.1    \\ \hline
FasterRCNN &  37.6 & 36.5/39.4    & 2.9  & 59.0 & 55.7/62.1 & 6.4  \\ \hline
CenterNet   &  34.6 & 32.9/36.3   & 3.4  & 53.0 & 49.2/57.3 & 8.1   \\ \hline
\end{tabular}
\caption{AP variations on the COCO validation set (the lower, the better), with a maximum shift of one pixel.}
\label{table:ap-variations-coco}
\end{table}

One other important result is that all methods are able to achieve better performance than the baseline by carefully selecting the best shifts for each validation image. This implies that shift variance could be leveraged to boost performance, as we show in a simple experiment in Section~\ref{sec:tta}.

\subsection{Increasing shift range}
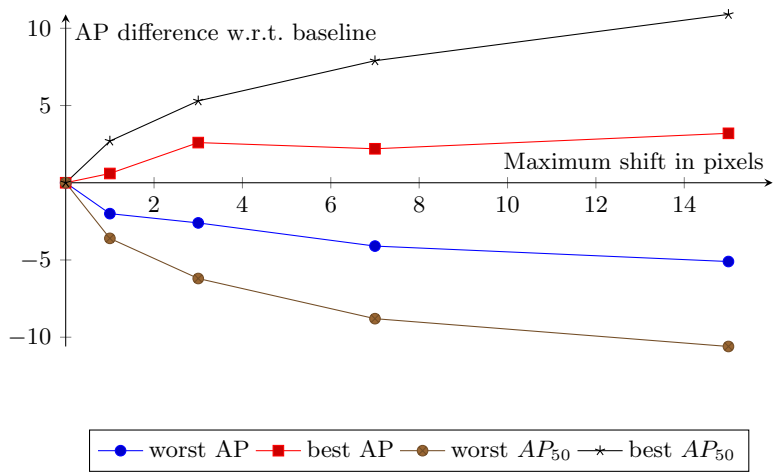
\begin{figure}[ht]
\centering
\begin{tikzpicture}
	\begin{axis}[
		axis lines = middle,
	    xlabel=Maximum shift in pixels,
		ylabel=AP difference w.r.t. baseline,
		height=6cm,
		width=0.9\linewidth,
		grid=minor,
		legend style={at={(.5,-0.25)},
		xmax=16,
		enlargelimits=false,
		xtickmax=15,
      anchor=north,legend columns=-1}]
	\addplot coordinates
		{(0,0) (1,-2) (3,-2.6) (7,-4.1) (15,-5.1)};
	\addplot coordinates
		{(0,0) (1,0.6) (3,2.6) (7,2.2) (15,3.2)};
	\addplot coordinates
		{(0,0) (1,-3.6) (3,-6.2) (7,-8.8) (15,-10.6)};
	\addplot coordinates
		{(0,0) (1,2.7) (3,5.3) (7,7.9) (15,10.9)};
	\legend{worst AP,best AP,worst $AP_{50}$,best $AP_{50}$}
	\end{axis}
\end{tikzpicture}
\caption{Difference between baseline results and best/worst APs computed by our method with increasing maximum shift.
Results are reported for RetinaNet on 100 random images from the COCO validation set.}
\label{fig:varying-shifts}
\end{figure}

We showed how a simple one-pixel shift can drastically affect the performance of several state-of-the-art detectors.
But how do the detectors behave for increasingly high shifts?
In this section we report the results of shift equivariance of RetinaNet with increasing maximum shift, from 0 to 15.
Since the number of images grows quadratically with the number of shifts (256 images for a maximum shift of 15), to save computation we perform the analysis on a randomly sampled validation set of 100 images. Results are reported in Table~\ref{table:varying-shift}, in Figure~\ref{fig:varying-shifts} we illustrate the AP difference with respect to the un-shifted baseline.
The AP variation grows sub-linearly with the maximum shift, reaching an impressive 21.5  $\Delta AP_{50}$ for a maximum shift of 15. The worst and best APs are symmetric with respect to the baseline AP, meaning that the baseline performance can be regarded as an \textit{average case} among the shifts. It can be noted that $\Delta AP_{50}$ grows monotonically with the maximum shift, but $\Delta AP$ does not. This is a result of our evaluation metric, that uses $AP_{50}$ to find best and worst shifts.

\begin{table}[]
\centering
\begin{tabular}{|l|l|l|l|l|}
\hline
\textbf{max shift} & $\mathbf{\sfrac{worst}{best}}$ $\mathbf{AP}$ & $\mathbf{\Delta AP}$ &$\mathbf{\sfrac{worst}{best}}$  $\mathbf{AP_{50}}$ &  $\mathbf{\Delta AP_{50}}$  \\ \hline
0 - baseline & 43.8       & -     & 64.2         & -    \\ \hline
1            & 41.8/43.4  & 1.6   & 60.7/66.9    & 6.2     \\ \hline
3            & 41.2/46.4  & 5.2   & 58.0/69.5    & 11.5     \\ \hline
7            & 39.6/46.0  & 6.4   & 55.4/72.1    & 16.7     \\ \hline
15           & 38.7/47.0  & 8.3   & 53.6/75.1    & 21.5     \\ \hline
\end{tabular}
\caption{AP variations of RetinaNet on 100 images from the COCO validation set. Results are reported for increasingly high shifts in pixels.}
\label{table:varying-shift}
\end{table}

\section{Dissecting translation equivariance}
So far, we showed how modern object detectors suffer from a lack of shift equivariance. A small shift in the input image can have huge effects on the output predictions.
In the literature, many attempts have been made to discover the root causes of the lack of shift invariance and to propose solutions to counter it.
To the best of our knowledge, no previous attempts have been made to evaluate how these can impact the performance of object detectors.

In the following, we dissect shift equivariance performance under the light of several factors that can potentially affect it.

\subsection{Anti-aliasing in downsampling layers}
We retrain RetinaNet and FasterRCNN using Blur Pool~\cite{zhang2019making} in their backbone network, ResNet-50. Blur Pool affects all downsampling layers, by limiting aliasing effects introduced by max pooling and strided convolutions. In practice we use a 3$\times$3 blur filter, as recommended in the original paper.
We report the results of $\Delta AP$ on the COCO validation set in Table~\ref{table:blur-pool}, compared to baseline methods.

\begin{table}[]
\centering
\begin{tabular}{|l|l|l|l|l|l|l|}
\hline
\textbf{method} & \textbf{AP} & $\mathbf{\sfrac{worst}{best}}$ $\mathbf{AP}$  & $\mathbf{\Delta AP}$ &$\mathbf{AP_{50}}$& $\mathbf{\sfrac{worst}{best}}$ $\mathbf{AP_{50}}$ & $\mathbf{\Delta AP_{50}}$ \\ \hline
RetinaNet   &  36.5 & 35.3/37.5   & 2.2 & 56.7 & 53.9/59.0 & 5.1    \\ \hline
RetinaNet+blurpool   &  35.2& 34.3/35.7   & 1.4 & 55.1  & 51.4/58.6 & 3.4    \\ \hline
FasterRCNN &  37.6 & 36.5/39.4   & 2.9 & 59.0 & 55.7/62.1 & 6.4   \\ \hline
FasterRCNN+blurpool &  37.8 & 36.6/38.6   & 2.0 & 58.7 & 56.3/60.9 & 4.6   \\ \hline
\end{tabular}
\caption{Effect of anti-aliased features on AP variations on the COCO validation set, with a maximum shift of one pixel.}
\label{table:blur-pool}
\end{table}

On the un-shifted validation set, blur pool slightly degrades performance for RetinaNet, while keeping FasterRCNN results basically unchanged.
Robustness to translation is improved for both methods in all metrics of about 30$\%$, suggesting that anti-aliased features are effective also for object detection. 
The results also confirm the reports from~\cite{azulay2019deep}, that anti-aliasing is only a partial solution to the problem, and different factors might be playing a role in shift equivariance. 

\subsection{Densifying the output space}

One possible reason to the lack of shift equivariance for object detection is the mismatch between input and output resolution. For example, the finest output resolution Retinanet can reach is eight times smaller compared to that of the input image. As for one-layer-output CenterNet, the output resolution is four times smaller. The mismatch results in discontinuities in the ground truth and anchor matching: moving even one pixel implies an anchor from being assigned from positive to negative and vice versa for a neighboring anchor. To investigate to what extent this can contribute to the shift variance, we densify the output space of CenterNet by adding two extra upsampling layers with transposed convolutions to match the output resolution to the input images. CenterNet is selected because it uses a single layer to make predictions instead of multiple layers as RetinaNet. 
Densifying the output space leads to two problems: i) the imbalance between foreground and background in the output space becomes even more severe given that increasing output resolution leads to 16 times more negative predictions and ii) the memory footprint due to a full-resolution output poses a problem in training efficiency. The only hyperparameter changes we apply to compensate for the higher resolution are: i) decreased $loss_{size}$ weight to 0.02 and ii) increased max pooling kernel size to 11 for NMS.
Dense CenterNet achieves an AP of 33.4 (baseline is 34.6).
However, we couldn't see many benefits for shift equivariance: the $ \Delta AP_{50}$ for the dense CenterNet is 7.6 (baseline 8.1) and the AP difference is 3.3 (baseline 3.4).

\subsection{Removing Photographer's Bias}
Photographer's bias is present in datasets composed of images taken by humans~\cite{azulay2019deep}. Humans tend to put objects at the center of the frame, and usually take pictures by standing vertically, almost perpendicular to the ground. Absolute location in the image frame becomes then a useful feature to recognize object categories (e.g. the sky is up, a sofa is on the floor, \dots).
As~\cite{kayhan2020translation} pointed out, it is possible for a CNN to exploit such biases, and to introduce shift variance as a mean to capture absolute object location.

\begin{figure}[ht]
\begin{subfigure}{.325\textwidth}
  \centering
  \includegraphics[width=.9\linewidth]{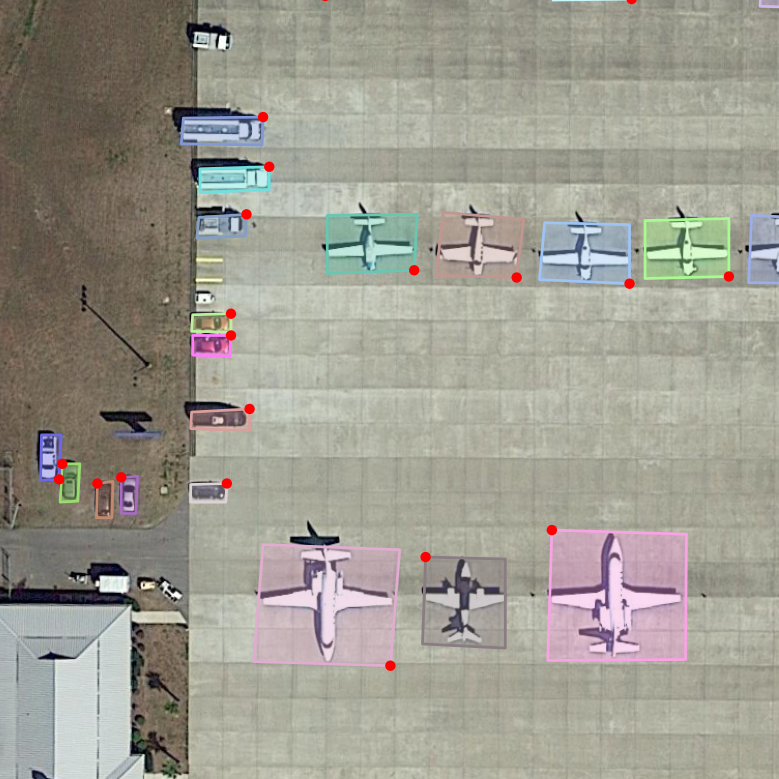}  
\end{subfigure}
\begin{subfigure}{.325\textwidth}
  \centering
  \includegraphics[width=.9\linewidth]{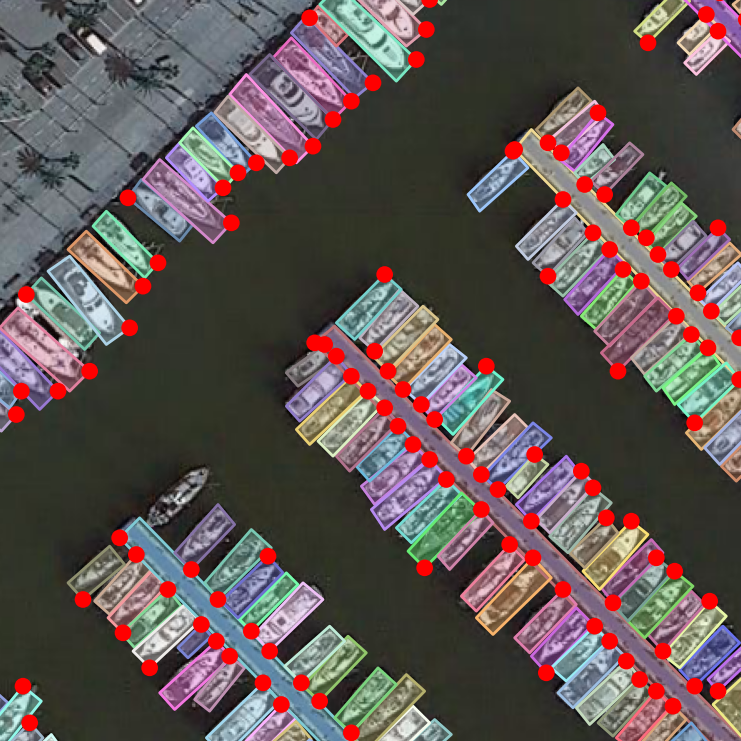}  
\end{subfigure}
\begin{subfigure}{.325\textwidth}
  \centering
  \includegraphics[width=.9\linewidth]{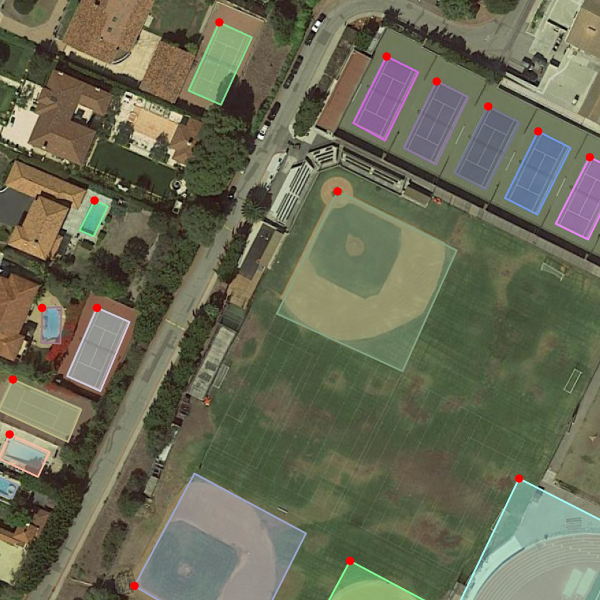}  
\end{subfigure}
\caption{Example crops from the DOTA object detection dataset. Instance annotation is available for rotated as for horizontal bounding boxes.}
\label{fig:dota-images}
\end{figure}

We therefore aim to measure the importance of the photographer's bias for shift equivariance in detection. To this end, we extended our quantitative analysis to the DOTA v1.0 object detection dataset~\cite{Xia_2018_CVPR}.
DOTA is the largest object detection dataset in aerial images, it contains 2806 large size images with $188,282$ object instances belonging to 15 categories, some sample crops are shown in Figure~\ref{fig:dota-images}. Aerial images do not suffer from the photographer's bias, since objects are not arranged in any particular order in the image frame.
The dataset defines two tasks: detection with oriented bounding boxes and detection with horizontal bounding boxes. We compute results for the latter, to keep the methods comparable to previous experiments.

\begin{table}[]
\centering
\begin{tabular}{|l|l|l|l|l|l|l|}
\hline
\textbf{method}       & $\mathbf{AP}$ & $\mathbf{\sfrac{worst}{best}}$ $\mathbf{AP}$  & $\mathbf{\Delta AP}$ & $\mathbf{AP_{50}}$  &$\mathbf{\sfrac{worst}{best}}$ $\mathbf{AP_{50}}$  & $\mathbf{\Delta AP_{50}}$ \\ \hline
RetinaNet   &  34.1 & 33.5/35.4   & 1.9 & 58.0 &55.7/60.5 & 4.8    \\ \hline
FasterRCNN &  36.7 & 36.0/38.7   & 2.7 & 60.3 & 57.4/63.2 & 5.8   \\ \hline
CenterNet &  34.6 & 33.5/35.7   & 2.2 & 57.3 & 54.4/60.2 & 5.8   \\ \hline
\end{tabular}
\caption{Shift equivariance evaluation on DOTA validation set, with a maximum shift of one pixel.}
\label{table:dota}
\end{table}

Following the experimental setting of~\cite{Xia_2018_CVPR}, images from the training set are split into patches of size 1024$\times$1024 with overlap of 512 pixels. This is necessary in order to fit the images to a CNN, since the original images have dimensions of 4000$\times$4000 pixels or more.
In the original experimental setting, validation images are also split into smaller crops, and the predictions from each crop are aggregated to compute the final performance. We retrained all detectors following the suggested setting and we obtained performance comparable with the results reported in~\cite{Xia_2018_CVPR}.

We evaluate shift equivariance on the validation crops, that can fit in memory for both the baseline and the shifted variants.
Shift equivariance results on the validation set, for a maximum shift of one pixel, are reported in Table~\ref{table:dota}.

Shift equivariance results are comparable to the ones of Table~\ref{table:ap-variations-coco}, however, all methods show an improvement compared to the COCO setting. CenterNet is the most different, with $\Delta AP$ going from 3.4 to 2.2 and $\Delta AP_{50}$ from 8.1 to 5.8.
Although an increase in robustness compared to the COCO dataset is measured, photographer's bias does not seem to be a major factor in translation equivariance.

\subsection{Test-Time Augmentation}
\label{sec:tta}
Test-time augmentation is commonly used to improve performance by computing predictions on several variations of the same image and then aggregating the predictions before evaluation. 
One example is multi-scale testing~\cite{singh2018analysis}, where each validation image is resized to different dimensions, predictions are computed for each scale and then mapped to a common reference frame, where non maximum suppression (NMS) is performed to aggregate them.

In the following, we perform test-time augmentation on the translated images. The predictions of each shift are mapped back to the original image coordinates, and aggregated using NMS. As for all other experiments, we use a maximum shift of one pixel, leading to 4 translated versions of each image. 

\begin{table}[]
\centering
\begin{tabular}{l|l|l|l|l|l|l|}
\cline{2-7}
                                  & \multicolumn{2}{l|}{\textbf{baseline}} & \multicolumn{2}{l|}{\textbf{w/ tta}} & \multicolumn{2}{l|}{\textbf{best APs}} \\ \hline
\multicolumn{1}{|l|}{\textbf{method}}      & $\mathbf{AP}$            & $\mathbf{AP_{50}}$          & $\mathbf{AP}$                 & $\mathbf{AP_{50}}$              & $\mathbf{AP}$                & $\mathbf{AP_{50}}$              \\ \hline
\multicolumn{1}{|l|}{RetinaNet}   & 36.5          & 56.7          & 36.7               & 56.9              & 37.5              & 59.0              \\ \hline
\multicolumn{1}{|l|}{FasterRCNN} & 37.6          & 59.0          & 38.4               & 59.5              & 39.4              & 62.1              \\ \hline
\multicolumn{1}{|l|}{CenterNet}   & 34.6          & 53.0          & 35.6               & 55.2              & 36.3              & 57.3              \\ \hline
\end{tabular}
\caption{Detection performance on COCO validation set. For each method we report baseline results on the un-shifted images, together with test-time augmentation results and best results obtained by the proposed method with maximum shift of one pixel.}
\label{table:tta}
\end{table}

We want to measure if test-time augmentation is capable of retaining the predictions from the best performing shifts, leading to better results than the baseline.
In this experiment we don't evaluate the shift equivariance, since all shifts are considered at the same time during aggregation. Results are reported in Table~\ref{table:tta}.

Test-time augmentation improves results for all methods, bringing them closer to the best achievable APs on the shifted validation set. For CenterNet, we report an improvement of 1.0 $AP$ and 2.2 $AP_{50}$ over the baseline.
Test-time augmentation is a very simple way of exploiting shift variance, and it comes with a considerable computational overhead (4 shifts per image, leading to 4$\times$ inference time).
We believe there is room for better ways of leveraging the lack of shift equivariance, that can turn out to be an exploitable property, instead of an undesirable side effect of modern CNNs.
\section{Conclusions}
We showed that object detection models exhibit considerable variation with even small translations applied to input images, which has been observed and reported in image classification tasks~\cite{azulay2019deep,engstrom2018rotation,zhang2019making}. In order to quantitatively evaluate shift equivariance of object detection models end-to-end, we proposed a greedy algorithm to search the lower and upper bounds of mean average precision on a test set. 
We found consistent large performance gaps on multiple modern object detection architectures for even a single-pixel shift. Apart from providing the evaluation pipeline, we investigated and demonstrated ways to mitigate this problem. 
With blur-pool, we improved the $\Delta_{AP}$ by around $30\%$ on both one-stage and two-stage detectors. 
By densifying the output space, we could improve shift equivariance by a small margin on $\Delta {AP_{50}}$. 
In addition, we observed improved robustness on the DOTA aerial imaging dataset, that doesn't suffer from the photographer's bias.
Furthermore, we showed that test-time augmentation could leverage the variance introduced by input translations and improve the network performance at the cost of significantly higher inference time.

However, none of the aforementioned solutions could remove shift variance completely.
Further investigations on architecture improvements, loss formulations or data augmentation techniques would shed more light on this topic.

\par\vfill\par

\clearpage
%
%
\bibliographystyle{splncs04}


\begin{thebibliography}{10}
\providecommand{\url}[1]{\texttt{#1}}
\providecommand{\urlprefix}{URL }
\providecommand{\doi}[1]{https://doi.org/#1}

\bibitem{agarwal2020towards}
Agarwal, V., Shetty, R., Fritz, M.: Towards causal vqa: Revealing and reducing
  spurious correlations by invariant and covariant semantic editing. In:
  Proceedings of the IEEE/CVF Conference on Computer Vision and Pattern
  Recognition. pp. 9690--9698 (2020)

\bibitem{Alcorn_2019_CVPR}
Alcorn, M.A., Li, Q., Gong, Z., Wang, C., Mai, L., Ku, W.S., Nguyen, A.: Strike
  (with) a pose: Neural networks are easily fooled by strange poses of familiar
  objects. In: Proceedings of the IEEE/CVF Conference on Computer Vision and
  Pattern Recognition (CVPR) (June 2019)

\bibitem{azulay2019deep}
Azulay, A., Weiss, Y.: Why do deep convolutional networks generalize so poorly
  to small image transformations? Journal of Machine Learning Research
  \textbf{20}(184),  1--25 (2019)

\bibitem{von2019simulating}
von Bernuth, A., Volk, G., Bringmann, O.: Simulating photo-realistic snow and
  fog on existing images for enhanced cnn training and evaluation. In: 2019
  IEEE Intelligent Transportation Systems Conference (ITSC). pp. 41--46. IEEE
  (2019)

\bibitem{cai2018cascade}
Cai, Z., Vasconcelos, N.: Cascade r-cnn: Delving into high quality object
  detection. In: Proceedings of the IEEE conference on computer vision and
  pattern recognition. pp. 6154--6162 (2018)

\bibitem{chen2017deeplab}
Chen, L.C., Papandreou, G., Kokkinos, I., Murphy, K., Yuille, A.L.: Deeplab:
  Semantic image segmentation with deep convolutional nets, atrous convolution,
  and fully connected crfs. IEEE transactions on pattern analysis and machine
  intelligence  \textbf{40}(4),  834--848 (2017)

\bibitem{dodge2016understanding}
Dodge, S., Karam, L.: Understanding how image quality affects deep neural
  networks. In: 2016 eighth international conference on quality of multimedia
  experience (QoMEX). pp.~1--6. IEEE (2016)

\bibitem{engstrom2018rotation}
Engstrom, L., Tsipras, D., Schmidt, L., Madry, A.: A rotation and a translation
  suffice: Fooling cnns with simple transformations. arXiv preprint
  arXiv:1712.02779  (2017)

\bibitem{fawzi2015manitest}
Fawzi, A., Frossard, P.: Manitest: Are classifiers really invariant? In:
  British Machine Vision Conference (BMVC). pp. 106.1--106.13 (2015)

\bibitem{goodfellow2014explaining}
Goodfellow, I.J., Shlens, J., Szegedy, C.: Explaining and harnessing
  adversarial examples. arXiv preprint arXiv:1412.6572  (2014)

\bibitem{gu2019using}
Gu, K., Yang, B., Ngiam, J., Le, Q., Shlens, J.: Using videos to evaluate image
  model robustness. arXiv preprint arXiv:1904.10076  (2019)

\bibitem{hendrycks2018benchmarking}
Hendrycks, D., Dietterich, T.G.: Benchmarking neural network robustness to
  common corruptions and surface variations. arXiv preprint arXiv:1807.01697
  (2018)

\bibitem{hendrycks2019augmix}
Hendrycks, D., Mu, N., Cubuk, E.D., Zoph, B., Gilmer, J., Lakshminarayanan, B.:
  Augmix: A simple data processing method to improve robustness and
  uncertainty. In: International Conference on Learning Representations (2019)

\bibitem{islam2019much}
Islam, M.A., Jia, S., Bruce, N.D.: How much position information do
  convolutional neural networks encode? In: International Conference on
  Learning Representations (2019)

\bibitem{kang2019testing}
Kang, D., Sun, Y., Hendrycks, D., Brown, T., Steinhardt, J.: Testing robustness
  against unforeseen adversaries. arXiv preprint arXiv:1908.08016  (2019)

\bibitem{kayhan2020translation}
Kayhan, O.S., Gemert, J.C.v.: On translation invariance in cnns: Convolutional
  layers can exploit absolute spatial location. In: Proceedings of the IEEE/CVF
  Conference on Computer Vision and Pattern Recognition. pp. 14274--14285
  (2020)

\bibitem{lin2017focal}
Lin, T.Y., Goyal, P., Girshick, R., He, K., Doll{\'a}r, P.: Focal loss for
  dense object detection. In: Proceedings of the IEEE international conference
  on computer vision. pp. 2980--2988 (2017)

\bibitem{lin2014microsoft}
Lin, T.Y., Maire, M., Belongie, S., Hays, J., Perona, P., Ramanan, D.,
  Doll{\'a}r, P., Zitnick, C.L.: Microsoft coco: Common objects in context. In:
  European conference on computer vision. pp. 740--755. Springer (2014)

\bibitem{michaelis2019benchmarking}
Michaelis, C., Mitzkus, B., Geirhos, R., Rusak, E., Bringmann, O., Ecker, A.S.,
  Bethge, M., Brendel, W.: Benchmarking robustness in object detection:
  Autonomous driving when winter is coming. arXiv preprint arXiv:1907.07484
  (2019)

\bibitem{ren2015faster}
Ren, S., He, K., Girshick, R., Sun, J.: Faster r-cnn: Towards real-time object
  detection with region proposal networks. In: Advances in neural information
  processing systems. pp. 91--99 (2015)

\bibitem{rosenfeld2018elephant}
Rosenfeld, A., Zemel, R.S., Tsotsos, J.K.: The elephant in the room. arXiv
  preprint arXiv:1808.03305  (2018)

\bibitem{singh2018analysis}
Singh, B., Davis, L.S.: An analysis of scale invariance in object detection
  snip. In: Proceedings of the IEEE conference on computer vision and pattern
  recognition. pp. 3578--3587 (2018)

\bibitem{sundaramoorthi2019translation}
Sundaramoorthi, G., Wang, T.E.: Translation insensitive cnns. arXiv preprint
  arXiv:1911.11238  (2019)

\bibitem{wu2019detectron2}
Wu, Y., Kirillov, A., Massa, F., Lo, W.Y., Girshick, R.: Detectron2.
  \url{https://github.com/facebookresearch/detectron2} (2019)

\bibitem{Xia_2018_CVPR}
Xia, G.S., Bai, X., Ding, J., Zhu, Z., Belongie, S., Luo, J., Datcu, M.,
  Pelillo, M., Zhang, L.: Dota: A large-scale dataset for object detection in
  aerial images. In: Proceedings of the IEEE Conference on Computer Vision and
  Pattern Recognition. pp. 3974--3983 (2018)

\bibitem{xie2017aggregated}
Xie, S., Girshick, R., Doll{\'a}r, P., Tu, Z., He, K.: Aggregated residual
  transformations for deep neural networks. In: Proceedings of the IEEE
  conference on computer vision and pattern recognition. pp. 1492--1500 (2017)

\bibitem{zhang2019towards}
Zhang, H., Wang, J.: Towards adversarially robust object detection. In:
  Proceedings of the IEEE International Conference on Computer Vision. pp.
  421--430 (2019)

\bibitem{zhang2019making}
Zhang, R.: Making convolutional networks shift-invariant again. In:
  International Conference on Machine Learning. pp. 7324--7334 (2019)

\bibitem{zhou2019objects}
Zhou, X., Wang, D., Kr{\"a}henb{\"u}hl, P.: Objects as points. In: arXiv
  preprint arXiv:1904.07850 (2019)

\end{thebibliography}
\end{document}